\documentclass[letterpaper]{article} 
\usepackage{aaai2026}  
\usepackage{times}  
\usepackage{helvet}  
\usepackage{courier}  
\usepackage[hyphens]{url}  
\usepackage{graphicx} 
\usepackage{natbib}  
\usepackage{caption} 
\usepackage{amssymb}   
\usepackage{mathrsfs}  

\usepackage{amsmath,amsfonts}
\usepackage{multirow}    
\usepackage{booktabs}    
\usepackage{array}       
\usepackage{graphicx}    
\usepackage{cellspace}   
\usepackage{threeparttable}
\usepackage{tabularx}
\usepackage{wasysym}
\newcolumntype{C}[1]{>{\centering\arraybackslash}p{#1}}
\usepackage{makecell} 

\usepackage{algorithm}
\usepackage{algpseudocode}

\usepackage{newfloat}
\usepackage{listings}
\DeclareCaptionStyle{ruled}{labelfont=normalfont,labelsep=colon,strut=off} 
\floatstyle{ruled}
\newfloat{listing}{tb}{lst}{}
\floatname{listing}{Listing}
\title{Radar-APLANC: Unsupervised Radar-based Heartbeat Sensing via Augmented 
Pseudo-Label and Noise Contrast}

\author{
	Ying Wang\textsuperscript{\rm 1}, Zhaodong Sun\textsuperscript{\rm 
	1}\footnote{Corresponding author.}, Xu Cheng\textsuperscript{\rm 1}, Zuxian 
	He\textsuperscript{\rm 1}, Xiaobai Li\textsuperscript{\rm 2}
}
\affiliations{
	\textsuperscript{\rm 1}Nanjing University of Information Science \& 
	Technology, Nanjing, China\\
	
	\textsuperscript{\rm 2}Zhejiang University, Hangzhou, China\\
	\{ying.wang, zhaodong.sun, xcheng, hezuxian\}@nuist.edu.cn, 
	xiaobai.li@zju.edu.cn
}

\usepackage{bibentry}

\begin{document}

\maketitle

\begin{abstract}
	Frequency Modulated Continuous Wave (FMCW) radars can measure subtle chest 
	wall oscillations to enable non-contact heartbeat sensing. However, 
	traditional radar-based heartbeat sensing methods face performance 
	degradation due to noise. Learning-based radar methods achieve better noise 
	robustness but require costly labeled signals for supervised training. To 
	overcome these limitations, we propose the first unsupervised framework for 
	radar-based heartbeat sensing via \textbf{A}ugmented 
	\textbf{P}seudo-\textbf{La}bel and \textbf{N}oise \textbf{C}ontrast 
	(Radar-APLANC). We propose to use both the heartbeat range and noise range 
	within the radar range matrix to construct the positive and negative 
	samples, respectively, for improved noise robustness. Our Noise-Contrastive 
	Triplet (NCT) loss only utilizes positive samples, negative samples, and 
	pseudo-label signals generated by the traditional radar method, thereby 
	avoiding dependence on expensive ground-truth physiological signals. We 
	further design a pseudo-label augmentation approach featuring adaptive 
	noise-aware label selection to improve pseudo-label signal quality. 
	Extensive experiments on the Equipleth dataset and our collected radar 
	dataset demonstrate that our unsupervised method achieves performance 
	comparable to state-of-the-art supervised methods. Our code, dataset, and 
	supplementary materials can be accessed from 
	\url{https://github.com/RadarHRSensing/Radar-APLANC}.
\end{abstract}

%
\begin{figure}[t]
	\centering
	\includegraphics[width=4\linewidth, height=7.5cm, 
	keepaspectratio]{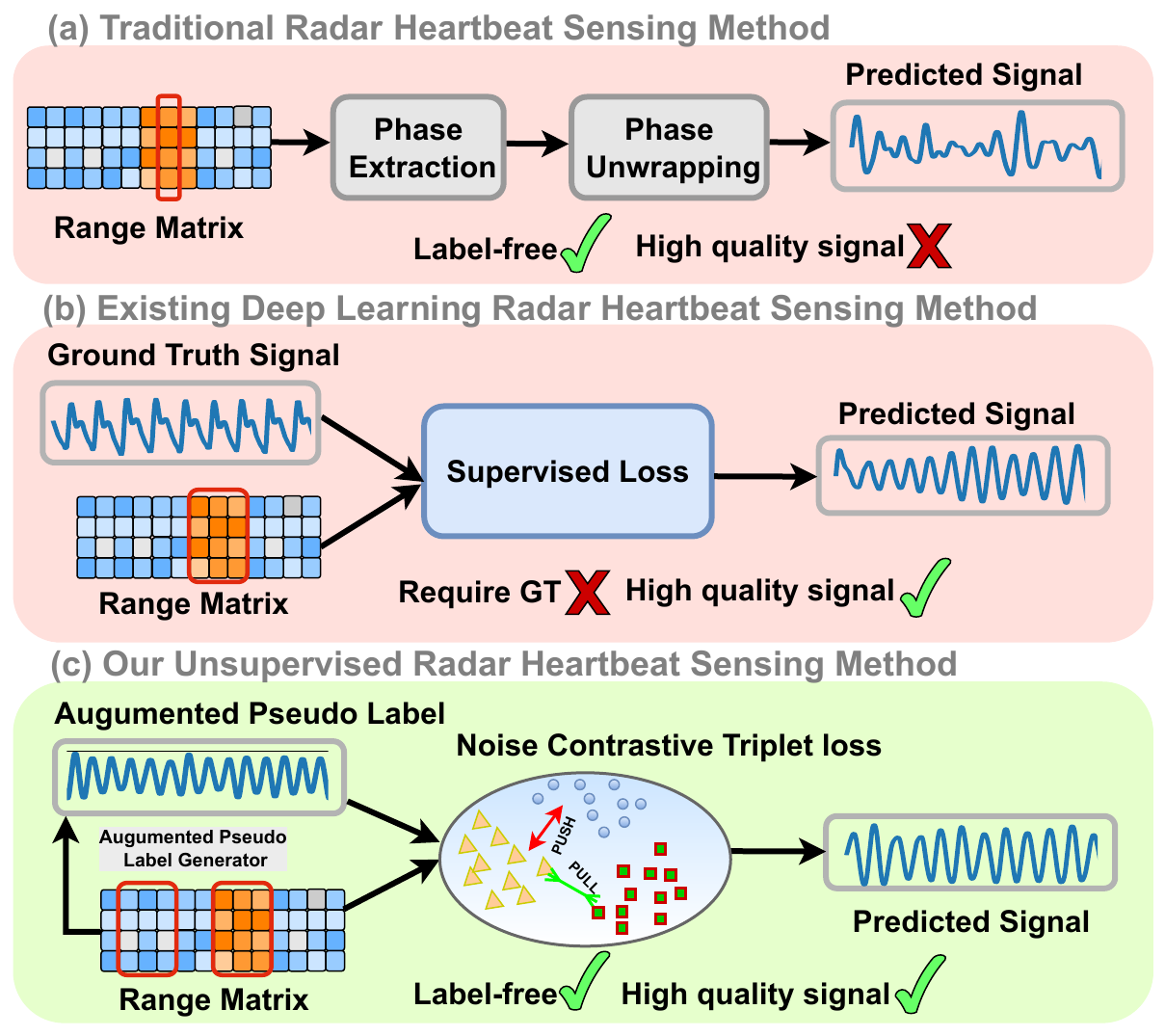}
	\caption{Comparison of radar-based heartbeat sensing methods. (a) 
	Traditional methods simply rely on phase extraction and unwrapping 
	processing, resulting in low signal quality; (b) Existing supervised 
	methods require ground truth signals; (c) Our proposed unsupervised 
	approach generates high-quality predictions without requiring ground truth 
	signals.}
	\label{fig:moti}
\end{figure}

\section{Introduction}
Radar-based heartbeat sensing has emerged as a pivotal technology for 
non-contact physiological assessment, offering distinct advantages in privacy 
preservation, environmental robustness, and continuous physiological 
monitoring. The basic principle for radar-based heartbeat sensing is detecting 
sub-millimeter chest wall displacements (typically 0.1–0.5 mm) induced by 
cardiac subtle motion \cite{droitcour2006non}. Frequency Modulated Continuous 
Wave (FMCW) radars are widely used to capture heartbeat signals by measuring 
the relative phase changes of the received chirp signals. Traditional 
approaches directly extract and unwrap the phases of received chirp signals 
\cite{alizadeh2019remote,7470533,mercuri2019vital}. However, these traditional 
methods suffer significant performance degradation under motion artifacts, 
multipath interference, and low signal-to-noise conditions due to inherent 
phase wrapping ambiguities and noise sensitivity as shown in Fig. 
\ref{fig:moti}(a).

Recent supervised deep learning methods 
\cite{vilesov2022blending,hu2024contactless,wu2025cardiacmamba} have 
demonstrated improved robustness by learning complex spatiotemporal patterns 
directly from radar data. However, these approaches require large-scale 
datasets with high-quality physiological annotations (e.g., synchronized PPG 
signals), which are costly to acquire, as shown in Fig. \ref{fig:moti}(b). This 
dependency creates the fundamental limitation for supervised radar methods: 
scalability bottlenecks of training data due to the expensive physiological 
annotations.

While unsupervised learning paradigms have shown promise in video-based 
physiological monitoring 
\cite{Gideon_2021_ICCV,sun2022contrast,speth2023non,li2023contactless,yue2023facial},
 directly adapting these to radar data faces inherent incompatibility and 
obstacles. Heartbeat signals in radar data usually exhibit lower 
signal-to-noise ratios compared to the video modality 
\cite{vilesov2022blending}; heartbeat signals manifest differently between 
radar versus video domains, i.e., chest motions for radar and facial color 
changes for videos; conventional positive and negative pair construction 
strategies (e.g., spatiotemporal similarity cross-sample dissimilarity) 
\cite{sun2022contrast} fail under strong noise interference in radar data. 
These challenges indicate that a specialized unsupervised framework should be 
designed for radar-based heartbeat sensing.

To address these challenges, we propose Radar-APLANC: a two-stage unsupervised 
radar-based heartbeat sensing framework built on augmented pseudo-label and 
noise contrast (APLANC) as shown in Fig. \ref{fig:moti}(c). Our approach 
introduces two key innovations: \textbf{1)} a Noise-Contrastive Triplet (NCT) 
loss is proposed by contrasting heartbeat signals with pseudo-label signals and 
noise signals from a radar range matrix; \textbf{2)} an augmented pseudo-label 
generator that refines pseudo-labels through quality assessment and adaptive 
noise-ware label selection. This two-stage framework enables effective 
unsupervised learning without physiological labels while maintaining noise 
robustness.

Comprehensive evaluations on the Equipleth dataset \cite{vilesov2022blending} 
and our collected radar heartbeat (RHB) dataset demonstrate that Radar-APLANC 
achieves close performance to existing state-of-the-art supervised methods. The 
results validate our unsupervised framework's capability to overcome 
fundamental limitations in traditional or supervised radar methods for 
heartbeat sensing.



The main contributions can be summarized as follows:

\begin{enumerate}
	\item We pioneer the first unsupervised framework for radar-based heartbeat 
	sensing, eliminating dependency on physiological labels and achieving 
	comparable performance with supervised methods.
	
	\item We propose the NCT loss – the first attempt to exploit noise 
	artifacts in radar range matrices to improve the noise robustness for 
	radar-based heartbeat sensing.
	
	\item We develop a two-stage training strategy with an augmented 
	pseudo-label generator featuring adaptive noise-aware label selection from 
	radar range matrices.
	
	\item We collected a new radar-based heartbeat sensing benchmark dataset 
	(RHB) from 80 subjects, which will be open-sourced for community research.
\end{enumerate}

\begin{figure*}[t]
	\centering
	\includegraphics[width=\textwidth]{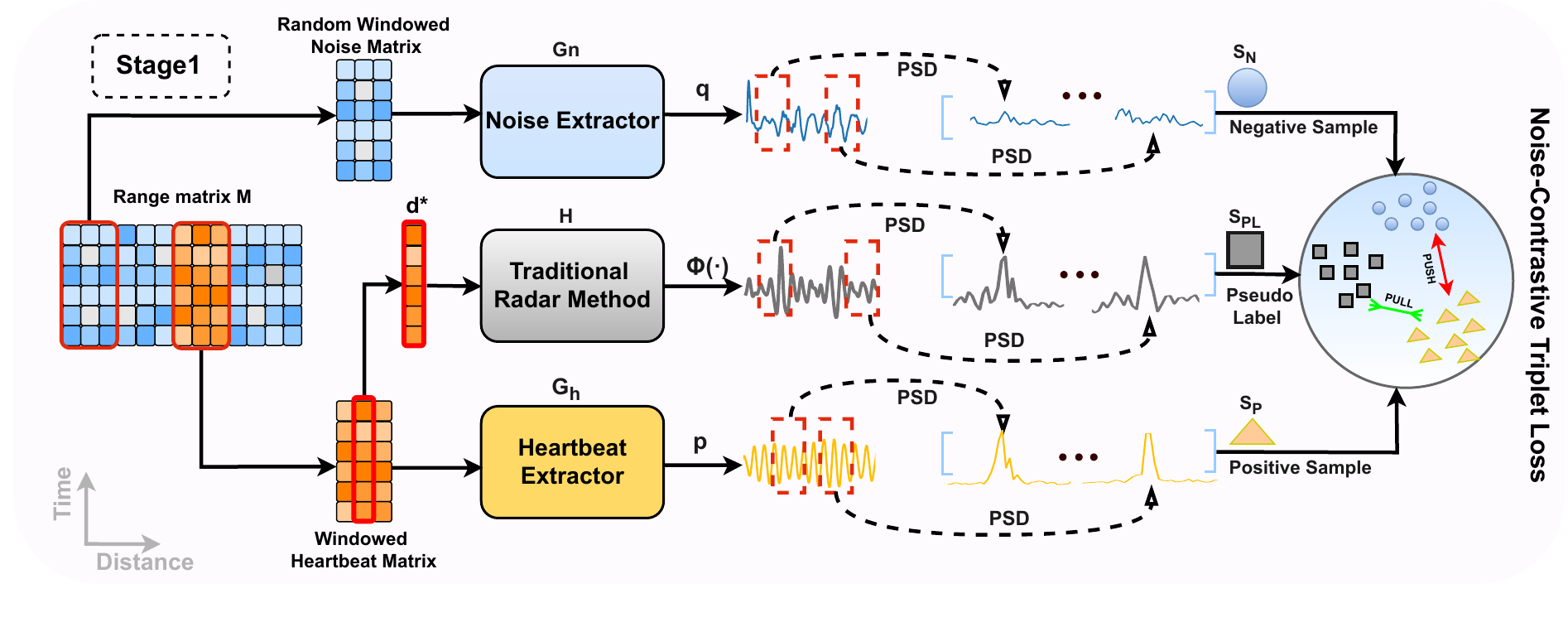} 
	\caption{The framework of Stage One. The heartbeat matrix, pseudo-label, 
	and random noise matrix undergo random temporal sampling and power spectrum 
	densities (PSD) transform before being fed into the NCT loss. Within this 
	framework, the PSD of the heart matrix is attracted to that of the 
	pseudo-label while being repelled from the noise PSD.}
	\label{fig1}
\end{figure*}

\begin{figure*}[t]
	\centering
	\includegraphics[width=\textwidth]{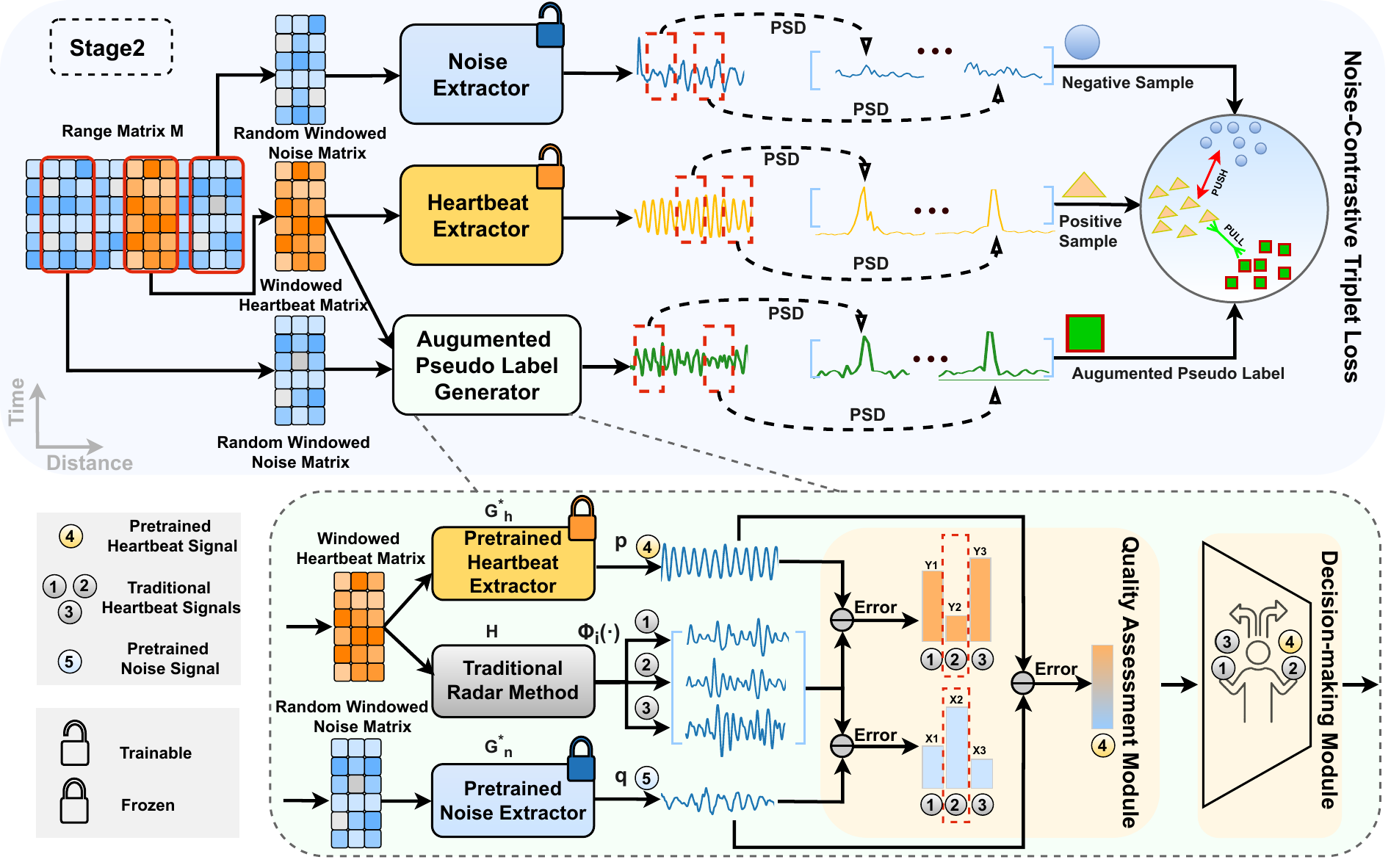} 
	\caption{The framework of Stage Two with Augmented Pseudo-label Generator. 
	It consists of Quality Measurement Module and Decision-making Module.}
	\label{fig2}
\end{figure*}

\section{Related Work}

\subsection{Radar-based Heartbeat Sensing}

Noncontact Radar-based heartbeat measurement has gained attention in recent 
works. \cite{alizadeh2019remote} combines phase unwrapping with range-FFT to 
detect heartbeat and respiration signals using FMCW radar in real-world 
settings. \cite{mercuri2019vital} introduces a radar system capable of 
simultaneously monitoring vital signs and tracking the spatial positions of 
multiple individuals without physical contact. \cite{7470533} presents a method 
that enables accurate respiration rate estimation under one-dimensional body 
motion using a single continuous-wave Doppler radar with motion compensation.

Recent efforts 
\cite{hu2024contactless,vilesov2022blending,wu2025cardiacmamba,10.1145/3550330} 
apply supervised deep learning to estimate heartbeat signals from radar data, 
achieving strong results by modeling temporal patterns. However, these methods 
require large annotated radar datasets, limiting training and data scalability.

To address this, recent 
studies~\cite{song2022rf,zhang2025umimo,song2024unleashing} explore 
self-supervised learning for radar-based sensing tasks, including human gesture 
recognition, 3D pose estimation, silhouette generation, and electrocardiography 
(ECG) signal reconstruction. While these studies reduce dependence on labeled 
data, they still require some annotations for fine-tuning and cannot achieve 
unsupervised learning.

\subsection{Unsupervised Learning with pseudo-labels}

Pseudo-labeling has been widely used in unsupervised learning, especially for 
classification. Early methods~\cite{saito2017asymmetric,chen2019progressive} 
improve label quality via multi-classifier refinement or progressive sample 
selection. Others leverage generative 
models~\cite{morerio2020generative,wang2022gpl} or contrastive 
strategies~\cite{sun2023contrastive,diamant2024confusing,litrico2023guiding} to 
address noisy labels in domain adaptation and source-free settings. Curriculum 
learning and debiasing approaches~\cite{choi2019pseudo,lai2023padclip} enhance 
pseudo-labeling by adjusting label difficulty and correcting imbalance during 
training.
In cross-modal tasks like visible-infrared person re-ID, clustering-based 
pseudo-labeling~\cite{shi2024multi} and soft 
assignment~\cite{Lin_2020_CVPR,seo2022unsupervised} further boost performance. 
However, these techniques are tailored for classification and are less 
applicable to regression tasks like heartbeat signal measurement. 
\cite{li2023contactless} utilizes traditional video-based methods to estimate 
heartbeat signals as pseudo-labels to achieve unsupervised training. Inspired 
by this work, our method initially generates pseudo-labels using traditional 
radar-based heartbeat estimation~\cite{alizadeh2019remote} as weak label 
supervision, but the noisy radar pseudo-labels are suboptimal. Therefore, we 
further incorporate noise information and design an augmented pseudo-label 
generator to adaptively select the radar pseudo-labels for improved performance.

\section{Method}
In this section, we will first introduce the preliminaries of radar-based 
heartbeat sensing as the basis for understanding our proposed method. 
Subsequently, we introduce our two-stage unsupervised method (Radar-APLANC) for 
radar-based heartbeat sensing. The overviews of the two stages are shown in 
Fig. \ref{fig1} and Fig. \ref{fig2}, respectively.

\subsection{Preliminaries}

\subsubsection{Range Matrix.}

A range matrix is obtained from FMCW radar raw data to facilitate the following 
analysis and processing. The procedures to obtain a range matrix are as 
follows. An FMCW radar in each chirp loop transmits a chirp signal $s(t)$ and 
receives the reflected chirp signal $u(t)$. Both  $s(t)$ and  $u(t)$ are linear 
frequency modulation signals, also called chirp signals. The received signal 
$u(t)$ is modulated with the in-phase and quadrature (IQ) transmitting signals 
$s_I(t)$ and $s_Q(t)$ to get the complex intermediate frequency (IF) signal 
$m(t)\in \mathbb{R}^{D}$ as shown below:
\begin{equation}
	\begin{aligned}
		m(t)  \propto &\text{LPF}[s_I(t) \cdot u(t)] + j \text{LPF}[s_Q(t) 
		\cdot u(t)]\\
		\propto & \exp(j(2\pi ft+\varphi)), f=2kd/c, \varphi=4\pi d/\lambda 
	\end{aligned}
	\label{eq:IF}
\end{equation}
where LPF is the low-pass filter, $k$ is the frequency slope of FMCW, $d$ is 
the distance, $c$ is the light speed, $\lambda$ is the wavelength of the FMCW 
starting frequency. The frequency $f$ of the IF signal $m(t)$ is the frequency 
difference between the transmitted signal $s(t)$ and the received signal 
$u(t)$. The frequency $f$ is also proportional to the signal round-trip time 
and the distance $d$ between the radar and the object. The phase $\varphi$ is 
also proportional to the distance but is bounded between $-\pi$ and $\pi$. A 
radar can sequentially transmit $N$ chirps $[s_1(t), s_2(t),..., s_N (t)]$ and 
receive the corresponding $N$ reflected chirps $[u_1(t), u_2(t),..., u_N (t)]$. 
In the meanwhile, the radar can obtain $N$ IF signals $[m_1(t), m_2(t),..., m_N 
(t)]$. Since the frequency of each IF signal $m_n(t)\in \mathbb{R}^{D}$ is 
related to the distance, the fast Fourier transform (FFT) is performed on each 
IF signal $m_n(t)$ to get the corresponding range profile $M_n[f]$. Finally, 
all range profiles are concatenated to obtain the range matrix:
\begin{equation}
	M=[M_1[f],M_2[f],...,M_N[f]] \in \mathbb{R}^{N \times D},
\end{equation}
where N is the number of chirps, and D is the number of range bins.

\subsubsection{Basic Radar-based Heartbeat Sensing.} The frequency $f$ cannot 
be directly used to extract the submillimeter chest motions caused by 
heartbeats since the range resolution is on the order of centimeters. Instead, 
the highly sensitive phase $\varphi$ should be used to measure such 
submillimeter displacement. The basic heartbeat sensing  
\cite{alizadeh2019remote} consists of four steps as shown in Fig. 
\ref{fig:moti}(a): (1) Select the range bin $d^*$ with the maximum power 
occupancy along the range axis in the range matrix. This range bin corresponds 
to the person's position. (2) Calculate the phase angle at the selected range 
bin for each range profile to get the phase signal 
$\phi(\cdot)=[\varphi_1,...,\varphi_N]=[\text{angle}(M_1[d^*]),...,\text{angle}(M_N[d^*])]
 \in \mathbb{R}^{N}$. (3) Since the phases are wrapped between $[-\pi,\pi]$, a 
standard phase unwrapping algorithm should be used to get the unwrapped phase 
signal. (4) Filter the unwrapped phase signal with 0.8 Hz-3.0 Hz bandpass to 
get the heartbeat signal $\Phi(\cdot) \in \mathbb{R}^{N}$ and find the highest 
peak corresponding to the heart rate in the frequency domain.

\subsection{Stage One: Unsupervised Noise-Contrastive Pretraining}
Fig. \ref{fig1} illustrates the framework of stage one, which leverages noise 
information and heartbeat information to construct pseudo-labels, positive 
samples, and negative samples for NCT loss. After stage one, our method can 
primarily extract coarse heartbeat signals.

\subsubsection{Pseudo-labels.} We use the traditional radar heartbeat sensing 
steps described above in the preliminaries section to get pseudo-labels. As 
shown in Fig. \ref{fig1} for Stage One, we use the traditional radar method 
\cite{alizadeh2019remote} to extract heartbeat signals from the range bin with 
the maximum power occupancy. We further perform random temporal sampling and 
power spectrum densities (PSD) transform shown in Fig. \ref{fig1} to generate 
multiple pseudo-labels for subsequent NCT Loss. The pseudo-label set $S_{PL}$ 
can be described as:
\begin{equation}
	S_{PL}=\{\mathcal{P}(\Phi(n_1\pm \Delta n)),...,\mathcal{P}(\Phi(n_K\pm 
	\Delta n))\},
\end{equation}
where $\mathcal{P}$ means PSD transform, $\Phi(\cdot) \in \mathbb{R}^{N}$ 
stands for the heartbeat signal from the traditional method, and $n_k \pm 
\Delta n$ represents the random time interval around $n_k$.

\subsubsection{Positive pairs.}Following the previous work 
\cite{vilesov2022blending}, we take a window of the range matrix $M\in 
\mathbb{R}^{N\times D}$ around the central range bin $d^*$ to get the windowed 
heartbeat matrix $M(\cdot, d^*\pm \Delta d) \in \mathbb{R}^{N\times (2\Delta 
d+1)}$ as the input to the heartbeat extractor. The extractor can output the 
predicted heartbeat signal $p(\cdot)\in \mathbb{R}^N$. Similar to 
pseudo-labels, we perform random temporal sampling and PSD transform on 
$p(\cdot)$ to get the positive sample set:
\begin{equation}
	S_P=\{\mathcal{P}(p(n_1\pm \Delta n)),...,\mathcal{P}(p(n_K\pm \Delta n))\}
\end{equation}
Since pseudo-labels contain heartbeat information, pseudo-labels and the 
predicted heartbeat signals should be similar, and the samples in the 
pseudo-label set $S_{PL}$ and the heartbeat signal set $S_P$ should be pulled 
together as positive pairs. Therefore, the positive term of the NCT loss can be 
represented as follows:
\begin{equation}
	\mathcal{L}_P=\frac{1}{K^2}\sum_{i=1}^{K}\sum_{j=1}^{K}||S_{PL}[i]-S_P[j]||^2
\end{equation}

\subsubsection{Negative pairs.} We form negative samples using random windowed 
noise matrices extracted from the range matrix. Since heartbeat information is 
primarily concentrated in the heartbeat matrix, other range bins mainly contain 
background noise. Therefore, we randomly select the range bin $d'$ except the 
central range bin $d^*$. Similar to the windowed heartbeat matrix, the windowed 
noise matrix is $M(\cdot, d'\pm\Delta d)$ as the input to the noise extractor. 
The noise extractor estimates the noise signal $q(\cdot)\in \mathbb{R}^N$. 
After temporal sampling and PSD transform, we get the negative sample set:
\begin{equation}
	S_N=\{\mathcal{P}(q(n_1\pm \Delta n)),...,\mathcal{P}(q(n_K\pm \Delta n))\}.
\end{equation}
Since heartbeat signals should not contain such background noises, the negative 
term of the NCT loss can be formulated as follows:
\begin{equation}
	\mathcal{L}_N=-\frac{1}{K^2}\sum_{i=1}^{K}\sum_{j=1}^{K}||S_{P}[i]-S_N[j]||^2
\end{equation}

The overall NCT loss is the sum of positive and negative loss terms: 
$\mathcal{L}_{NCT}=\mathcal{L}_P+\mathcal{L}_N$. The detailed training 
procedures of stage one are presented in Algorithm \ref{alg:algorithm}.

\begin{algorithm}[t!]
	\caption{Radar-APLANC Training Algorithm}
	\label{alg:algorithm}
	\textbf{Input}: Range matrix dataset $\{M_1,...M_Q\}$; Heartbeat extractor 
	$G_h$; Noise extractor $G_n$; Pretrained heartbeat extractor (only for 
	stage two) $G_h^*$; Pretrained noise extractor (only for stage two) 
	$G_n^*$; Traditional radar method $H$; Training stage STAGE.
	\begin{algorithmic}[1] 
		
		\For{e = 1:NumIteration}
		\State Load a range matrix $M_i$.
		
		\State \textcolor{gray}{\textit{For the positive sample}}
		\State Find the central range bin $d^*$ in $M_i$.
		\State $p=G_h(M_i(\cdot, d^*\pm \Delta d))$ \Comment{predicted 
		heartbeat signal}
		
		\State \textcolor{gray}{\textit{For the negative sample}}
		\State Randomly select a range bin $d'$ except $d^*$ in $M_i$.
		\State $q=G_n(M_i(\cdot,d'\pm \Delta d))$ \Comment{noise signal}
		
		\State \textcolor{gray}{\textit{For the pseudo-label}}
		\If {STAGE==1}
		\State $\Phi=H(M_i(\cdot, d^*))$ \Comment{traditional method}
		\ElsIf {STAGE==2}
		\State  $\Phi$ = \Call{AugPseudoGen}{$M_i$, $d^*$, $G_h^*$, $G_n^*$}
		\EndIf
		\State Calculate NCT loss and update $G_h$ and $G_n$.
		\EndFor
		\State \textcolor{gray}{\textit{Augmented Pseudo-label Generator}}
		\Function{AugPseudoGen}{$M$, $d^*$, $G_h^*$, $G_n^*$}
		\State $\Phi_1, ..., \Phi_{2\Delta d+1}=H(M(\cdot, d^*\pm \Delta d))$ 
		\Comment{traditional heartbeat}
		\State $p=G_h^*(M(\cdot, d^*\pm \Delta d))$ \Comment{pretrained 
		heartbeat}
		\State Randomly select a range bin $d'$ except $d^*$ in $M$.
		\State $q=G_n^*(M(\cdot, d'\pm \Delta d))$ \Comment{pretrained noise}
		\State \textcolor{gray}{\textit{Noise distance $X_i$ and heartbeat 
		distance $Y_i$}}
		\State $X_i, Y_i = D(\Phi_i, q), D(\Phi_i, p), i\in[1, 2\Delta d+1]$ 
		\State \textcolor{gray}{\textit{Decision-making Module}}
		\If{$\arg \max_i X_i==\arg \min_i Y_i$}
		\State Return $\Phi_{\arg \min_i Y_i}$
		\Else
		\If{$X_{\arg \min_i Y_i}>D(p,q)$} 
		\State Return $\Phi_{\arg \min_i Y_i}$
		\Else
		\State Return $p$
		\EndIf
		\EndIf
		\EndFunction
	\end{algorithmic}
\end{algorithm}

\subsection{Stage Two: Unsupervised Learning with Augmented Pseudo-label 
Generation}

Due to the significant noise inherent in radar data, the performance in stage 
one, using the traditional radar method to generate pseudo-labels, remains 
limited. Therefore, we introduce an augmented pseudo-label generator to further 
refine and optimize the selection of pseudo-labels as illustrated in Figure 
\ref{fig2}.


\subsubsection{Candidate Signal Extraction.} We extract a pretrained heartbeat 
signal $p$ from the windowed heartbeat matrix $M(\cdot, d^*\pm\Delta d)$ and a 
pretrained noise signal $q$ from a random windowed noise matrix $M(\cdot, 
d'\pm\Delta d)$ using the stage one pretrained extractors. Simultaneously, we 
extract $2\Delta d+1$ traditional heartbeat signals $\{\Phi_1, ..., 
\Phi_{2\Delta d+1}\}$ from the windowed heartbeat matrix $M(\cdot, d^*\pm\Delta 
d)$ using the traditional radar method. The final pseudo-label will be selected 
from the traditional heartbeat signals $\{\Phi_1, ..., \Phi_{2\Delta d+1}\}$ 
and the pretrained heartbeat signal $p$, while the noise signal $q$ will be 
used to assess the heartbeat signal qualities.

\subsubsection{Quality Assessment Module.} We compute two types of distances 
for candidate heartbeat signals $\{\Phi_1, ..., \Phi_{2\Delta d+1}\}$: (1) 
Noise distance $X_i$ between candidate heartbeat signals $\{\Phi_1, ..., 
\Phi_{2\Delta d+1}\}$ and the pretrained noise signal $q$: 
\begin{equation}
	X_i = D(\Phi_i, q), i \in [1, 2\Delta d+1].
\end{equation}
Longer noise distances indicate better signal quality. (2) Heartbeat distance 
$Y_i$  between  the traditional heartbeat signals $\{\Phi_1, ..., 
\Phi_{2\Delta+1}\}$ and the pretrained heartbeat signal $p$:
\begin{equation}
	Y_i = D(\Phi_i, p), i \in [1, 2\Delta d+1].
\end{equation}
which measures how similar the traditional heartbeat signal $\Phi_i$ is to the 
pretrained heartbeat signal $p$. Since we have an assumption that the 
pretrained heartbeat signal $p$ has good signal quality, shorter distances 
indicate better signal quality. To measure the distance between two signals, we 
use the mean absolute error between the two signals' heart rates.


\subsubsection{Decision-making Module.} The best heartbeat signal among 
$\{\Phi_1, ..., \Phi_{2\Delta+1}\}$ should satisfy the condition that its noise 
distance $X_i$ should be maximum while its heartbeat distance $Y_i$ should be 
minumum. Therefore, the following equation should hold: $\arg \max_i X_i=\arg 
\min_i Y_i$. When this ideal condition is not met, i.e., $\arg \max_i X_i \ne 
\arg \min_i Y_i$, the module evaluates whether the traditional heartbeat signal 
$\Phi_{\arg \min Y_i}$ with the minimum heartbeat distance is sufficiently 
distant from the noise $q$. Specifically, it checks if $\Phi_{\arg \min Y_i}$'s 
noise distance $X_{\arg \min Y_i}$ is greater than the pretrained heartbeat 
signal's noise distance $D(p,q)$. If yes, the signal quality of $\Phi_{\arg 
\min Y_i}$ is better than that of $p$, and the module will select $\Phi_{\arg 
\min Y_i}$. Otherwise, the module will select the pretrained heartbeat signal 
$p$. This selection strategy can adaptively select the best-quality heartbeat 
signal as the enhanced pseudo-label. The detailed training procedures of stage 
two are presented in Algorithm \ref{alg:algorithm}.

\section{Experiments}

\begin{table*}[htbp]
	\centering
	\setlength{\tabcolsep}{3pt}
	\resizebox{\linewidth}{!}{
		\begin{tabular}{@{}  l ccc ccc ccc ccc @{}}
			\toprule
			\multirow{2}{*}{\textbf{Method}} &
			\multicolumn{3}{c}{\textbf{Equipleth}} &
			\multicolumn{3}{c}{\textbf{RHB$\rightarrow$Equipleth}} & 
			\multicolumn{3}{c}{\textbf{RHB}}&
			\multicolumn{3}{c}{\textbf{Equipleth$\rightarrow$ RHB}} \\
			\cmidrule(lr){2-4} \cmidrule(lr){5-7} \cmidrule(lr){8-10} 
			\cmidrule(l){11-13}
			& \textbf{MAE}↓ & \textbf{RMSE}↓ & \textbf{r}↑ & 
			\textbf{MAE}↓ & \textbf{RMSE}↓ & \textbf{r}↑ & 
			\textbf{MAE}↓ & \textbf{RMSE}↓ & \textbf{r}↑ & 
			\textbf{MAE}↓ & \textbf{RMSE}↓ & \textbf{r}↑ \\
			\midrule
			
			\begin{tabular}[c]{@{}l@{}}FFT-based RF 
			\cite{alizadeh2019remote}$\blacktriangle$\end{tabular} 
			& 13.51 & 21.07 & 0.24 & 13.51 & 21.07 & 0.24 & 12.25 & 18.37 & 
			0.26& 12.25 & 18.37 & 0.26  \\
			
			\begin{tabular}[c]{@{}l@{}}\cite{7470533}$\blacktriangle$\end{tabular}
			 
			& 5.50 & 11.68 & 0.64 & 5.50 & 11.68 & 0.64 & - & - & - & - & - & - 
			\\
			\midrule
			\begin{tabular}[c]{@{}l@{}}Equipleth RF 
			\cite{vilesov2022blending}$\blacklozenge$\end{tabular} 
			& \textbf{2.18} & \textbf{6.12} & \textbf{0.89} & \underline{4.53} 
			& 9.63 & 0.65& \textbf{3.19} & \textbf{7.18} & \textbf{0.82} & 
			\underline{2.68} & \underline{6.29} & \underline{0.86}  \\
			
			%
			%
			
			\begin{tabular}[c]{@{}l@{}}mmFormer 
			\cite{hu2024contactless}$\blacklozenge$\end{tabular} 
			& 6.50 & 11.10 & 0.52 & 7.72 & 11.73 & 0.40 & 8.89 & 12.77 & 0.28& 
			7.00 & 7.83 & 0.47  \\
			
			\begin{tabular}[c]{@{}l@{}}VitaNet (Khan et al. 
			2022)$\blacklozenge$\end{tabular} 
			& \underline{3.14} & \underline{7.70} & \underline{0.77} & 7.43 & 
			11.86 & 0.40 & 5.28 & 9.25 & 0.66& \textbf{2.38} & \textbf{5.14} & 
			\textbf{0.90}  \\
			\midrule  
			\begin{tabular}[c]{@{}l@{}}Radar-APLANC 
			(ours)$\bigstar$\end{tabular} 
			& 3.95 & 9.72 & 0.64 & \textbf{4.10} & \textbf{8.51} & 
			\textbf{0.72}& \underline{3.92} & \underline{7.94} & 
			\underline{0.77} & 3.52 & 7.45 & 0.79  \\
			\bottomrule
	\end{tabular}}
	
	\medskip
	\begin{tablenotes}
		\item $\blacktriangle$: Traditional Training-free Methods,
		$\blacklozenge$: Supervised Methods,
		$\bigstar$: Unsupervised Methods
	\end{tablenotes}
	\caption{Intra-dataset and cross-dataset heart rate results of radar 
	modality on Equipleth dataset and our RHB dataset. The best results are in 
	bold, and the second-best results are underlined.}
	\label{tab:1}
\end{table*}

\begin{table*}[t]
	\centering
	\setlength{\tabcolsep}{15pt}
	\normalsize
	\resizebox{\linewidth}{!}{
		\begin{tabular}{@{}l l ccc @{}}
			\toprule
			\multirow{3}{*}{\textbf{Method}} & 
			\multirow{3}{*}{\textbf{Modality}} & 
			\multicolumn{3}{c}{\textbf{Performance (Fairness)}} \\
			\cmidrule(lr){3-5} 
			& & \textbf{MAE↓(→0)(bpm)} & \textbf{RMSE↓(→0)(bpm)} & 
			\textbf{r↑(→0)} \\ 
			
			\midrule
			CHROM \cite{de2013robust}  $\blacktriangle$ & \multirow{4}{*}{RGB} 
			& 7.45(4.97) & 13.38(4.17)  & 0.46(-0.38) \\
			ICA \cite{poh2010advancements}$\blacktriangle$ & & 8.38(4.42) & 
			14.03(3.15)  & 0.41(-0.36) \\
			PhysNet \cite{yu2019remote}  $\blacklozenge$   & & 1.78(2.22) & 
			5.26(4.05)  & 0.91(-0.25) \\
			FusionPhys-RGB \cite{ying2025} $\blacklozenge$   & & 1.49(1.23) & 
			5.53(3.50)  & 0.89(-0.12) \\
			\midrule
			
			FFT-based RF \cite{alizadeh2019remote}$\blacktriangle$ & 
			\multirow{5}{*}{Radar} & 13.51(2.25)  & 21.07(2.47)   & 
			0.240(-0.25) \\
			VitaNet (Khan et al. 2022)$\blacklozenge$   &  & 
			3.14(\textbf{0.30}) & 7.70(\textbf{0.48})  & 0.77(\textbf{-0.04}) \\
			mmFormer \cite{hu2024contactless}  $\blacklozenge$   &  & 
			6.50(\underline{0.34}) & 11.10(3.13)  & 0.52(-0.28) \\
			Equipleth RF \cite{vilesov2022blending}  $\blacklozenge$   &  & 
			2.18(0.51) & 6.12(\underline{0.85})  & 0.89(-0.13) \\
			Radar-APLANC (ours)$\bigstar$   &  & 3.95(0.91) & 9.72(0.98) & 
			0.64(\underline{-0.06})   \\
			\bottomrule
	\end{tabular}}
	\medskip
	\begin{tablenotes}
		\item $\blacktriangle$: Traditional Training-free Methods,  
		$\blacklozenge$ Supervised Methods, $\bigstar$: Unsupervised Methods
	\end{tablenotes}
	\caption{Heart rate performance and skin tone fairness on Equipleth dataset 
	among radar methods and RGB methods. Among the radar methods, the best 
	results are in bold, and the second-best results are underlined.}
	\label{tab:comparison}
\end{table*}

\subsection{Datasets and Implementation}

\subsubsection{Equipleth Dataset.} The public radar dataset 
\cite{vilesov2022blending} contains 550 paired facial video and FMCW radar 
recordings from 91 subjects. Skin tones are categorized by the Fitzpatrick 
scale \cite{sachdeva2009fitzpatrick}: 28 light, 49 medium, and 14 dark for skin 
tone fairness evaluation. Each subject contributed six 30-second recordings. 
More details are presented in the supplementary materials.


\subsubsection{RHB Dataset.} Our collected RHB dataset comprises 240 instances 
of FMCW radar data from 80 volunteers. Each participant has three separate 
30-second recording sessions while maintaining a seated position approximately 
0.5 to 1 meter in front of the radar acquisition board, consistent with 
\cite{vilesov2022blending}. Data was captured at 120 frames per second. The 
detailed dataset configuration is presented in the supplementary materials.



\subsubsection{Implementation Details.}


Following prior work \cite{vilesov2022blending}, we use 10-second windows for 
training and heart rate evaluation. For the Equipleth dataset, we use the same 
training protocol as \cite{vilesov2022blending}. For our RHR dataset, we apply 
4-fold cross-validation. Each fold contains 50 subjects for training, 10 for 
validation, and 20 for testing.
Both the noise and heartbeat extractors are randomly initialized and trained in 
two stages. The augmented pseudo-label generator inherits pretrained models 
from stage one. Each phase is trained with AdamW (learning rate 1e-4) for 200 
epochs, and the best epoch is chosen by validation sets. The heart rate 
evaluation follows prior work using mean absolute error (MAE), root mean 
squared error (RMSE), and Pearson correlation (r).


\subsection{Comparsions with state-of-the-art methods}

\subsubsection{Intra-dataset testing.}
Table \ref{tab:1} presents the performance of our unsupervised method on the 
Equipleth and our proposed RHB datasets. On the Equipleth dataset, supervised 
methods lead—Equipleth RF method \cite{vilesov2022blending} and VitaNet 
\cite{10.1145/3550330} rank first and second, respectively. Impressively, our 
unsupervised method achieves comparable accuracy, with only 25.8\% higher MAE 
than VitaNet. On RHB dataset, our approach achieves near state-of-the-art 
performance (MAE=3.92), just 22.9\% above the Equipleth RF method. Notably, our 
method shows strong stability on the two datasets, whereas other supervised 
models degrade significantly on RHB dataset. In addition, our unsupervised 
method also significantly outperforms traditional training-free radar methods. 
These results highlight the stability and robustness of Radar-APLANC.


\subsubsection{Cross-dataset testing.} Our unsupervised method demonstrates 
strong generalization across datasets. When trained on RHB and tested on 
Equipleth (RHB→Equipleth), it incurs only a +3.8\% MAE increase, significantly 
outperforming supervised methods (Equipleth Radar: +107.8\%, VitaNet: 
+136.6\%). In the reverse setting (Equipleth→RHB), all methods improve, yet our 
method demonstrates superior stability, exhibiting minimal performance 
deviation between direct training-testing on RHB and cross-dataset transfer 
from Equipleth to RHB. Notably, our MAE fluctuation across intra- and 
cross-dataset evaluations remains within 0.4 bpm. These results validate the 
robustness of our framework in learning domain-invariant representations, while 
supervised approaches exhibit unstable cross-domain transfers.



\begin{figure}[t]
	\centering
	\includegraphics[width=\linewidth, keepaspectratio]{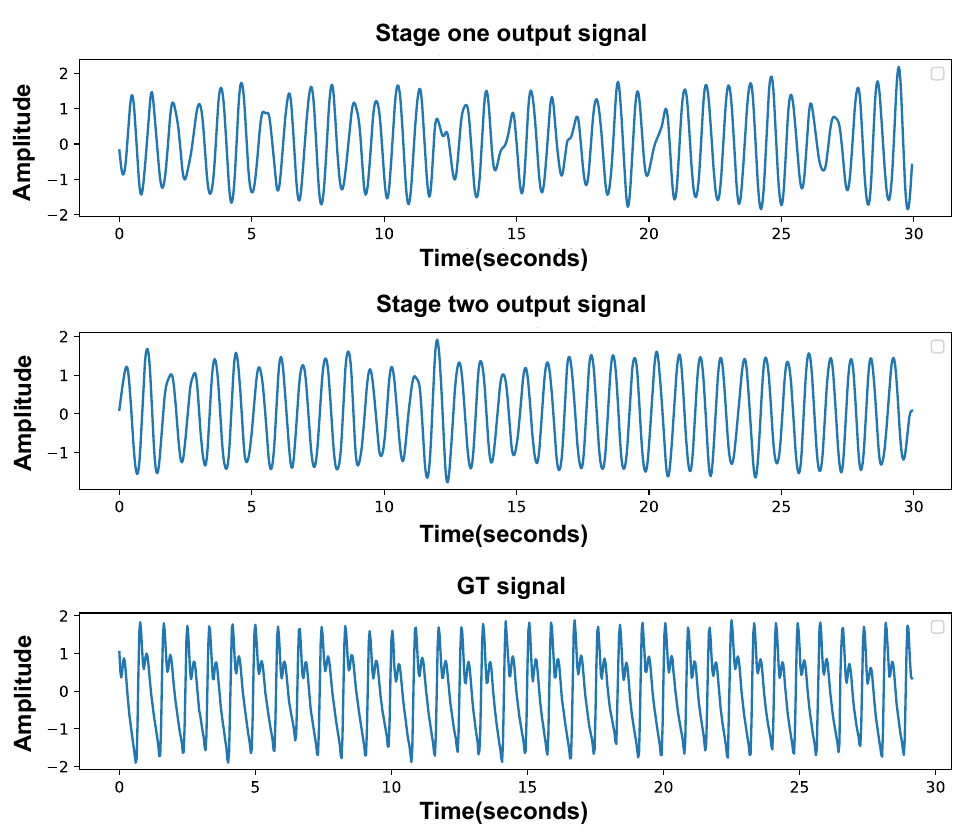}
	\caption{Example heart pulse signals generated by our stage one and stage 
	two models and the ground truth signal.}
	\label{fig:example}
\end{figure}

\subsection{Performance and Fairness Comparison with Other Modalities}

Table~\ref{tab:comparison} compares the performance and fairness across sensing 
modalities in the EquiPleth dataset. Fairness is the performance difference 
between dark and light skin tones, and being close to zero indicates high 
fairness. RGB-based methods—especially supervised ones (e.g., 
PhysNet~\cite{yu2019remote})—achieve high overall performance but exhibit low 
fairness, indicating significant performance drops on dark skin tones due to 
low reflected light from dark skin. In contrast, radar-based approaches 
generally demonstrate better fairness than RGB since radar is independent of 
skin tones and lighting conditions. These findings align with prior 
work~\cite{vilesov2022blending} that RGB methods are more prone to skin tone 
bias than radar-based ones. Notably, our unsupervised radar method exhibits 
comparable fairness with other radar methods and outperforms RGB methods in 
terms of fairness, highlighting the potential of label-free radar-based sensing 
for equitable heartbeat monitoring.

\subsection{Ablation Study and Visualization}

Table \ref{tab:ablation} presents the ablation study results of our method on 
the Equipleth dataset. When only pseudo-labels are utilized in stage one, our 
approach achieves an MAE of 8.94, surpassing the traditional FFT-based RF 
method (MAE 13.51) by a significant margin \cite{alizadeh2019remote}. While the 
noise matrix alone fails to converge (MAE 34.48), its combination with 
pseudo-labels substantially reduces the MAE to 4.4, less than half of the 8.94 
using only pseudo-labels. This demonstrates the feasibility of effectively 
leveraging both noise matrices and pseudo-labels. Furthermore, our augmented 
pseudo-label approach in stage two consistently outperforms all configurations 
in stage one. Specifically, when all configurations in both stages are enabled, 
it achieves the lowest MAE of 3.95. These results collectively validate the 
efficacy of our augmented pseudo-labeling strategy and noise matrices. 

Table \ref{tab:ablation2} presents the ablation study of the augmented 
pseudo-label generator in stage two. It can be observed that traditional 
heartbeat signals combined with the pretrained heartbeat signal or the noise 
signal for signal quality assessment and decision-making yield poor performance 
(MAE 8.75 and 14.48). Although directly employing the pretrained heartbeat 
signal as pseudo labels yields better results than the former two, it (MAE 
4.56) falls short compared to the first-stage performance (MAE 4.40) shown in 
Table \ref{tab:ablation}, thereby diminishing the value of label augmentation. 
By combining all three types of signals, the best performance is achieved.

\begin{table}[t]
	\centering
	\resizebox{\linewidth}{!}{
		\begin{tabular}{@{} ccccccc @{}}
			\toprule
			\multicolumn{2}{c}{\textbf{Stage 1}} & 
			\multicolumn{2}{c}{\textbf{Stage 2}} &
			\multicolumn{3}{c}{\textbf{Metrics}}\\
			\cmidrule(lr){1-2} \cmidrule(lr){3-4}  \cmidrule(lr){5-7} 
			\textbf{\makecell{Noise \\ Matrix}} & 
			\textbf{\makecell{Pseudo \\ label}} & 
			\textbf{\makecell{Augumented \\ Pseudo \\ label}} & 
			\textbf{\makecell{Noise \\ Matrix}} &
			\textbf{\makecell{MAE↓}} & 
			\textbf{\makecell{RMSE↓}} & 
			\textbf{\makecell{r↑}} \\
			\midrule
			$\checkmark$&  &  & &34.48 & 38.34& 0.01\\
			& $\checkmark$ &  & &8.94 & 15.88& 0.30\\
			$\checkmark$& $\checkmark$ &  & &4.40 & 9.89& 0.63\\
			\midrule
			$\checkmark$&$\checkmark$& $\checkmark$ && 7.42 & 13.61& 0.38\\
			$\checkmark$& $\checkmark$ &$\checkmark$ &$\checkmark$& 
			\textbf{3.95} & \textbf{9.72}& \textbf{0.64}\\
			\bottomrule
	\end{tabular}}
	\caption{Ablation study of the critical components in two stages on 
	EquiPleth dataset. The best result is in bold.}
	\label{tab:ablation}
\end{table}

\begin{table}[t]
	\centering
	\resizebox{\linewidth}{!}{
		\begin{tabular}{@{}c c c c c c@{}}
			\toprule
			\multicolumn{1}{c}{\begin{tabular}[c]{@{}c@{}}\textbf{Pretrained} 
			\\ \textbf{Heartbeat} \\\textbf{Signal}\end{tabular}} & 
			\multicolumn{1}{c}{\begin{tabular}[c]{@{}c@{}}\textbf{Traditional} 
			\\ \textbf{Heartbeat}\\ \textbf{Signals}\end{tabular}} &
			\multicolumn{1}{c}{\begin{tabular}[c]{@{}c@{}}\textbf{Noise} \\ 
			\textbf{Signal} \end{tabular}} &
			\multicolumn{1}{c}{\textbf{MAE↓}} & 
			\multicolumn{1}{c}{\textbf{RMSE↓}} & 
			\multicolumn{1}{c}{\textbf{r↑}} \\ 
			\midrule $\checkmark$&  & & 4.56 & 10.09 &0.63 \\ 
			$\checkmark$&  $\checkmark$& & 8.75& 15.09 & 0.30\\
			&  $\checkmark$& $\checkmark$& 14.48& 17.76 & 0.13\\
			$\checkmark$& $\checkmark$ &$\checkmark$ & \textbf{3.95} & 
			\textbf{9.72} & \textbf{0.64}\\ 
			\bottomrule
	\end{tabular}}
	\caption{Ablation study on Augmented Pseudo Label Generator on EquiPleth 
	dataset. The best results are in bold.}
	\label{tab:ablation2}
\end{table}

As shown in Figure~\ref{fig:example}, stage one already produces heartbeat 
signals closely aligned with the ground truth signal, with only minor 
deviations in some segments. Stage two further refines these outputs, 
effectively reducing residual fluctuations and achieving improved waveform 
accuracy. The visual result in stage one demonstrates the effectiveness of NCT 
loss and pseudo-label in the radar unsupervised learning. In addition, the 
visual result in stage two reveals that augmented pseudo-labels can further 
improve heartbeat signals. The visualization results are also consistent with 
the ablation study in Table \ref{tab:ablation}.


\section{Conclusion}
This paper presents Radar-APLANC, the first unsupervised framework for 
radar-based heartbeat sensing. To eliminate reliance on physiological signal 
labels and improve noise robustness, we introduce a novel NCT loss that 
contrasts predicted heartbeat signals with pseudo-labels and noise signals. We 
further design an augmented pseudo-label generator to improve the pseudo-label 
quality to further improve the unsupervised learning performance. Unsupervised 
Radar-APLANC is comprehensively validated to be comparable with other 
supervised methods on the public Equipleth dataset and our collected RHB 
dataset.

\section{Acknowledgments}
This work was supported by the National Natural Science Foundation of China 
(Grant No. 62572249), the Natural Science Foundation of Jiangsu Province (Grant 
No. BK20250742), and the Startup Foundation for Introducing Talent of NUIST 
(Grant No. 1083142501006).

\bibliography{aaai2026}

\end{document}